Author's version (arXiv v2) of the peer-reviewed article published open access in *Artificial Intelligence in Health* (AccScience Publishing). Received June 19, 2026; revised July 21, 2026; accepted August 4, 2026; published online September 7, 2026.  

# Automated multi-class wound assessment using dedicated instance segmentation models for boundary detection and classification

Mehedi Hasan Tusar[1,*], Fateme Fayyazbakhsh[1,2,3], Igor Melnychuk[4], and Ming C. Leu[1,3,5]

[1] Department of Mechanical and Aerospace Engineering, College of Engineering and Computing, Missouri University of Science and Technology, Rolla, Missouri, United States of America
[2] Missouri Protoplex, Missouri University of Science and Technology, Rolla, Missouri, United States of America
[3] Center for Biomedical Research, Missouri University of Science and Technology, Rolla, Missouri, United States of America
[4] Wound Care Department, Charles George Department of Veterans Affairs Medical Center, Asheville, North Carolina, United States of America
[5] Intelligent Systems Center, College of Engineering and Computing, Missouri University of Science and Technology, Rolla, Missouri, United States of America

*__Corresponding author:__ Mehedi Hasan Tusar (meheditusar95@gmail.com)

## Abstract

Accurate wound classification (WC) and boundary segmentation are essential for guiding clinical decisions in both chronic and acute wound management. However, most existing artificial intelligence (AI) models are limited, focusing on a narrow set of wound types, limited variations in wound severity, or a single task (segmentation or classification), which reduces their clinical applicability. This study presents two dedicated instance segmentation models based on You Only Look Once (YOLO)v11 that perform wound boundary segmentation (WBS) and WC across five clinically relevant wound types: burn injury (BI), pressure injury, diabetic foot ulcer, vascular ulcer, and surgical wound. A wound-type balanced dataset of 2,963 annotated images was created to train the models for both tasks, using five-fold cross-validation to ensure robust and unbiased evaluation. Models trained on the original, non-augmented dataset achieved consistent performance across folds, though BI detection accuracy was relatively low. Therefore, the dataset was augmented using rotation, flipping, and variations in brightness, saturation, and exposure to help the model learn more generalizable, invariant features. This augmentation significantly improved performance, particularly for visually subtle BI cases. Among the tested variants, YOLOv11x achieved the best WBS performance (F1-score: 0.9341; mean average precision [mAP]50: 0.9629). For WC, YOLOv11m achieved the highest mAP50 (0.9194) and mAP50–95 (0.6950), whereas YOLOv11l achieved the highest F1-score (0.8797). The lightweight YOLOv11n provided comparable accuracy at lower computational cost, making it suitable for resource-constrained deployments. Supported by confusion matrices and visual detection outputs, the results confirm the model's robustness against complex backgrounds and high intra-class variability, demonstrating the potential of YOLOv11-based architectures for accurate, real-time wound analysis in both clinical and remote care settings.

# 1. Introduction

A chronic wound is a wound that fails to progress through the normal stages of healing in an expected timeframe, usually persisting for more than six weeks. Chronic wounds are often associated with underlying conditions such as diabetes, poor circulation, pressure, or infection. Common types include diabetic foot ulcers (DFUs), pressure injuries (PIs), and vascular ulcers (VUs), such as venous leg ulcers and arterial ulcers.[1,2] Chronic wounds are among the most pressing challenges in modern healthcare, affecting millions of patients worldwide and imposing a significant socioeconomic burden. In the United States alone, chronic wounds impact more than six million individuals annually, with the total cost of wound care estimated between USD $28.1 and USD $96.8 billion.[3,4] Within United States Medicare expenditures, surgical wounds (SWs) represent the largest category among wounds, with costs ranging from USD $11.7 billion to USD $38.3 billion, including infection management.[4] Burn injuries (BIs) further add to this burden, accounting for 111,292 deaths worldwide in 2019 and approximately 327,000 new cases in the United States, with direct medical costs exceeding USD $976.6 million and indirect costs surpassing USD $1.17 billion.[5,6] As of 2024, the global wound care product market was valued between USD $15 billion and USD $22 billion, reflecting the demand for accurate, efficient, and standardized wound diagnosis and management tools.[7]

Advances in artificial intelligence (AI), particularly deep learning (DL), offer new opportunities to automate wound diagnosis. Two critical components of wound care diagnosis are wound boundary segmentation (WBS), which refers to the precise localization of wound margins, and wound classification (WC), which involves identifying the wound type (e.g., DFU, BI, PI).[8–12] These are essential for estimating wound size, monitoring healing trajectory, selecting appropriate treatment plans, and evaluating intervention efficacy.[13,14] In current clinical practice, however, assessment typically relies on visual inspection and manual tools such as rulers, tracing sheets, or photographs.[15,16] These methods are subjective, prone to inter-observer variability, and often slow, particularly in busy or resource-limited settings.[17] Convolutional neural networks (CNNs), a form of DL in particular, can learn hierarchical features from wound images, enabling accurate pixel-level segmentation and wound type classification.[18,19] These capabilities reduce the time, labor, and subjectivity of manual assessment while providing reproducible, large-scale diagnostic support. By delivering standardized, real-time insights, intelligent systems could assist general practitioners, nurses, and patients before specialist consultation, enabling timely and more consistent care. Accurate wound assessment can shorten healing time by supporting appropriate treatment, objective care planning, and reliable monitoring of healing progression. It can also strengthen telemedicine by enabling remote evaluation in settings with limited access to specialists. To be clinically viable, however, such systems must perform robustly across diverse wound presentations and imaging conditions while remaining computationally efficient for deployment in point-of-care (POC) and mobile settings.[20,21]

Recent progress in DL has greatly advanced automated wound assessment, especially in WBS and WC. Early segmentation studies often treated wound boundary detection and tissue classification as separate tasks, which limited their clinical use. To overcome this, Sarp *et al*.[22] introduced a conditional generative adversarial network that performed both tasks together, achieving high accuracy on over 2,000 annotated wound images without extra post-processing. Chang *et al*.[23] and Chauhan and Goyal[24] further improved BI segmentation using networks such as DeepLabV3+, Mask R-CNN, and ASPP-ResNet101, reporting accuracies above 90%. Niri *et al*.[25] proposed a U-Net model with dual attention blocks that captured key spatial and feature details, achieving a Dice score of 94.1%. Shah *et al*.[26] developed CADFU, a combined detection and segmentation framework using You Only Look Once (YOLO)v8 and DHuNeT, which showed high precision and robustness. Pandey *et al*.[27] explored thermal imaging-based segmentation for non-invasive PI diagnosis using SSD MobileNetV2, while Li *et al*.[28] improved surgical healing assessment by combining traditional image preprocessing with a MobileNet-based segmentation approach of thermal imaging data. To address limited labeled data, Dhar *et al*.[29] proposed a semi-supervised model using a transformer-based encoder and lightweight decoder, achieving consistent results even with small datasets for DFU segmentation. Similarly, Scebba *et al*.[30] presented a Detect-and-Segment pipeline that maintained high accuracy even when training data was reduced. Together, these studies show that hybrid, attention-based, and semi-supervised DL models improve segmentation quality and make wound analysis more adaptable to clinical challenges.

Research on WC has followed a similar evolution, moving from simple binary tasks to more advanced multi-class and multi-modal approaches. Huang *et al*.[31] developed a CNN to classify five wound types as deep, infected, arterial, venous, and pressure while achieving performance comparable to human experts. Moses *et al*.[32] later suggested that using pretrained models such as ResNet and Inception improved accuracy and training efficiency. Rostami *et al*.[3] introduced an ensemble CNN that combined patch-level and full-image analysis, reaching over 90% accuracy for different wound types. For BI, Yadav *et al*.[33] presented BuRnGANeXt50, a deep residual network with attention mechanisms, achieving sensitivities above 97%. More recently, Patel *et al*.[34] and Anisuzzaman *et al*.[35] developed multi-modal systems that combined wound images with anatomical location data using architectures such as VGG16, ResNet152, and EfficientNet. These approaches achieved up to 100% accuracy and clearly outperformed models relying on images alone. Overall, current DL-based methods show a clear shift toward unified, data-efficient, and multi-modal frameworks that enhance clinical reliability and improve wound diagnosis and monitoring.

Despite the potential of DL, several limitations persist across existing wound analysis studies. Many models are trained on relatively small, imbalanced datasets, which limit generalizability and increase the risk of overfitting, particularly in diverse real-world settings.[3,36,37] Wound images sometimes exhibit high intra-class variability and low inter-class separability—different wound types may appear visually similar, especially with dressings, occlusions, or inconsistent lighting—posing a challenge for robust classification.[29,33,38] Most classification models rely solely on image data, omitting valuable contextual features such as anatomical location or patient history, thereby limiting clinical applicability.[34,35] A critical gap in the literature is the lack of wound-type diversity in training datasets—many high-performance WBS models are trained on only one or two categories, such as DFU or PI, and often neglect BI, VU, and SW.[30] WC models often exclude certain wound types or include them in an imbalanced way, leading DL models trained on limited categories to generalize poorly and misclassify novel or underrepresented wounds.[39,40] In addition, many existing models are computationally demanding, making them impractical for deployment on POC devices in mobile or resource-limited settings. Critically, most studies also lack validation on external datasets, direct comparisons with human experts, or evaluation in real-world clinical workflows, which hinders their acceptance in routine wound care.[31,41] Addressing these challenges is essential for developing AI systems that are accurate, generalizable, and clinically relevant.

Because whole-image classifiers do not localize the wound, they are prone to missed detections and misclassification for small or low-contrast wounds; localizing the wound before labeling mitigates this across the severity range in our dataset. To address these gaps, we prepared an instance segmentation-based DL system that performs WBS and multi-class WC across all five major wound types: BI, PI, DFU, VU, and SW. To the best of our knowledge, no prior work has addressed both tasks across five major wound categories using the same wound images. The key contributions of this work are:

(i) Development of comprehensive wound datasets that include both WBS and WC across five major categories.

(ii) Expansion of the datasets through augmentation techniques including flipping, rotation, and variations in brightness, exposure, and saturation to enhance the model's ability to learn robust and invariant features under diverse lighting and geometric conditions.

(iii) Preparing and fine-tuning a DL model based on the pretrained YOLOv11 architecture, enabling WBS and WC.

(iv) Comparative evaluation of YOLOv11 variants—nano (n), small (s), medium (m), large (l), and extra-large (x)—to assess diagnostic performance and identify optimal trade-offs between accuracy and computational efficiency.

(v) Visual inference testing and benchmarking of the prepared models against existing classification and instance segmentation approaches for wound type classification, demonstrating the system's effectiveness and generalizability.

In the remainder of this paper, the methodology section describes the study design, dataset preparation, model architecture, and evaluation process. The results and discussion section presents outcomes from both the original and augmented datasets, performance comparisons among YOLOv11 variants, and benchmarking against other models. Finally, the conclusion section summarizes key findings and provides recommendations for future work.

## 2. Methods

### 2.1. Study design and dataset preparation

This study was designed as a retrospective diagnostic analysis and adhered to the Standards for Reporting of Diagnostic Accuracy Studies 2015 guidelines to ensure comprehensive and transparent reporting of diagnostic performance evaluation.[42,43] A retrospective diagnostic analysis looks backward in time, using existing datasets that have already been collected and labeled to evaluate the diagnostic accuracy of a test, algorithm, or model.

We initially collected 3,500 wound images from publicly available sources, including GitHub, Roboflow, Kaggle, and general internet searches. After excluding images with poor resolution, excessive brightness, or other quality issues, 2,963 high-quality, de-identified images were finalized. Because the models perform instance segmentation, multiple wounds within a single image were detected and segmented individually; each wound instance was annotated separately with its own mask and class label. In the dataset, 571 of 2,963 images (19.3%) contained more than one wound instance, all of the same wound type. These images were consolidated from the above sources and re-annotated under a single consistent protocol; the complete prepared dataset is openly available (see Availability of Data). These images represented five major wound types—PI, BI, DFU, VU, and SW—and some had previously undergone clinical validation.[29,34,35,39] As all images were de-identified and publicly accessible, no ethical or institutional review board approval was required.

Both WBS and WC tasks used the same pool of 2,963 annotated wound images but were organized into two independent, task-specific, non-augmented datasets. The WBS dataset was designed exclusively for boundary detection, while the WC dataset focused on wound type identification. During dataset construction, Roboflow's built-in duplicate detection was applied to automatically identify and remove exact and near-duplicate images prior to fold splitting, reducing the risk of data leakage from repeated images across the aggregated sources. All images were auto-oriented and resized to 640 × 640 pixels prior to annotation using Roboflow.[44] This resolution is the standard input size for the YOLO architecture and was selected to balance wound-detail preservation with memory and computational efficiency; it also matched the predominantly small-to-medium wound sizes in our dataset and supported real-time inference. Data augmentation was applied separately for each task, including horizontal and vertical flips, 90° rotations, and controlled variations in saturation (±25%), brightness (±15%), and exposure (±10%). The non-augmented WBS and WC datasets, each containing 2,963 images, were expanded to 7,147 images in their augmented versions by applying the described augmentation techniques. These augmentations were applied only to the training portion of each fold, after fold splitting; validation images were left unaltered.

The WBS dataset contained a single class ("wound") and was used to train the instance segmentation model. The WC dataset was annotated across five wound classes—PI, BI, DFU, VU, and SW—representing varying wound etiologies and severities. Although both datasets originated from the same clinical image pool, they served distinct model objectives: the WBS dataset for pixel-level boundary detection and the WC dataset for instance-level wound type classification.

Both tasks were trained as independent YOLOv11 instance-segmentation models using an identical hyperparameter configuration (Table 1), ensuring the models differed only in task and training data rather than in optimization settings. Image annotation was performed by a single trained annotator under the direct supervision of two senior wound-care experts, each with more than 15 years of experience. Both experts were present during the annotation process and reviewed and verified each wound-type label and boundary delineation to establish an expert-validated consensus ground truth. Because the source images were collected with clearly specified wound types and most wounds had boundaries readily distinguishable from the surrounding skin, annotations could be assigned consistently. As annotation followed this supervised single-consensus protocol rather than independent parallel annotation by multiple raters, a formal inter-rater agreement statistic (e.g., Cohen's kappa) was not applicable; instead, agreement was ensured through direct expert oversight and verification of every annotation. A schematic overview of the entire workflow is illustrated in Figure 1. All model training was performed in the Missouri University of Science and Technology's high-performance computing cluster, Mill, using an NVIDIA V100 (32 GB) graphics processing unit.

**Table 1.** Training hyperparameters used for all YOLOv11 variants across both the wound boundary segmentation and wound classification tasks

| Parameter | Value |
|---|---|
| Task | Instance segmentation |
| Pretrained weights | YOLOv11{n,s,m,l,x}-seg |
| Epochs | 150 |
| Image size (pixels) | 640 × 640 |
| Batch size (images per batch) | 16 or 32 |
| Optimizer | auto (SGD-based) |
| Initial LR (lr0) | 0.01 |
| Final LR factor (lrf) | 0.01 |
| Momentum | 0.937 |
| Weight decay | 0.0005 |
| Warmup epochs | 3 |
| Box/Cls/DFL weights | 7.5 / 0.5 / 1.5 |
| Mask ratio/overlap | 4/True |
| IoU threshold (val, NMS) | 0.7 |
| Max detections | 300 |
| Mixed precision (AMP) | Enabled |

Abbreviations: AMP: Automatic mixed precision; DFL: Distribution focal loss; IoU: Intersection over union; LR: Learning rate; NMS: Non-maximum suppression; SGD: Stochastic gradient descent; YOLO: You Only Look Once.

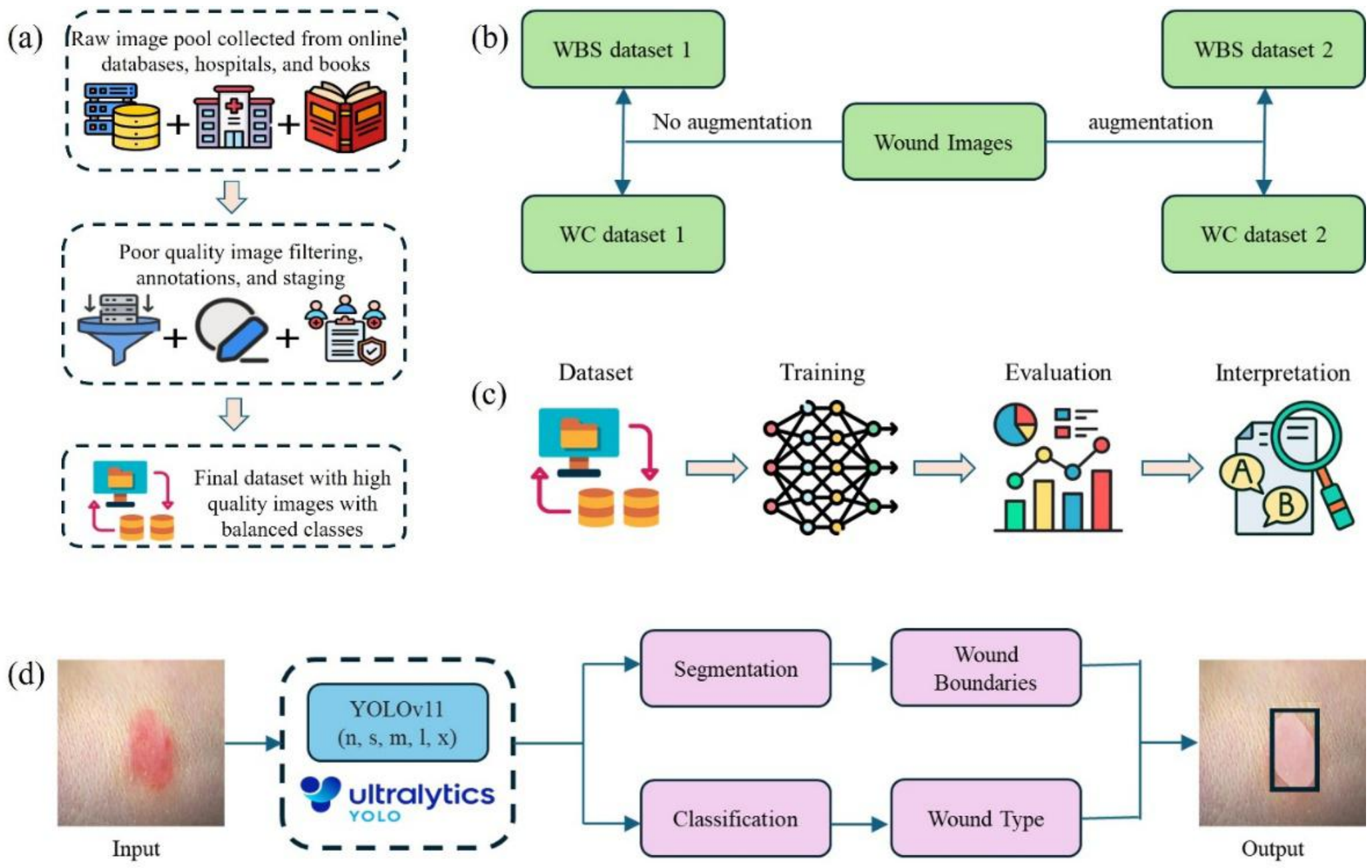

**Figure 1.** Workflow of the proposed system illustrating (a) data collection, (b) dataset generation, (c) model training and evaluation pipeline, and (d) inference outputs for wound boundary segmentation (WBS) and wound classification (WC).

## 2.2. Model architecture

YOLOv11, released by Ultralytics in October 2024, is a state-of-the-art model for real-time object detection and instance segmentation that builds upon the YOLOv8 framework.[45] Designed to balance computational efficiency and prediction accuracy, YOLOv11 is well-suited for high-resolution real-time medical imaging tasks.[46] As illustrated in Figure 2, the architecture follows a backbone–neck–head design that enables hierarchical feature extraction, multi-scale feature fusion, and multi-task prediction.[47] YOLOv11 replaces several C2f modules used in YOLOv8 with C3k2 modules, which are efficient C2f-derived cross-stage partial bottlenecks with two main convolutional layers and configurable internal blocks. These C3k2 blocks enable efficient multi-scale feature extraction with fewer parameters—YOLOv11m, for example, reduces parameter count by 22% compared to YOLOv8m—while preserving representational power via channel separation and residual connections.[45,48] The neck comprises a spatial pyramid pooling-fast module and a path aggregation network, facilitating effective fusion of semantic and spatial features at multiple scales with reduced computational overhead.[49,50] For prediction, the head is composed of two branches: an anchor-free detection head that regresses bounding box coordinates and class probabilities, and a segmentation head that leverages a prototype-based approach to generate instance-level masks.[39,45] To enhance focus on significant regions, the model integrates a cross-stage partial spatial attention, which prioritizes localized features and improves segmentation accuracy.[51] The output includes bounding boxes, classification scores, and pixel-level instance masks, enabling WC and boundary delineation. The model was trained using a compound loss that combines binary cross-entropy and intersection over union (IoU)-based metrics for segmentation, cross-entropy for classification, and auxiliary terms such as distribution focal loss to ensure stable convergence across heterogeneous data, while the task-aligned assigner matches predictions with ground-truth instances during training.[52] Post-processing steps included non-maximum suppression to eliminate redundant detections and prototype mask decoding to construct clean, per-instance boundaries. Collectively, YOLOv11's innovations—C3k2 blocks for lightweight feature learning, spatial pyramid pooling-fast for efficient receptive field expansion, and cross-stage partial spatial attention for spatial attention—make it a robust architecture for real-time segmentation and classification applications.[53,54]

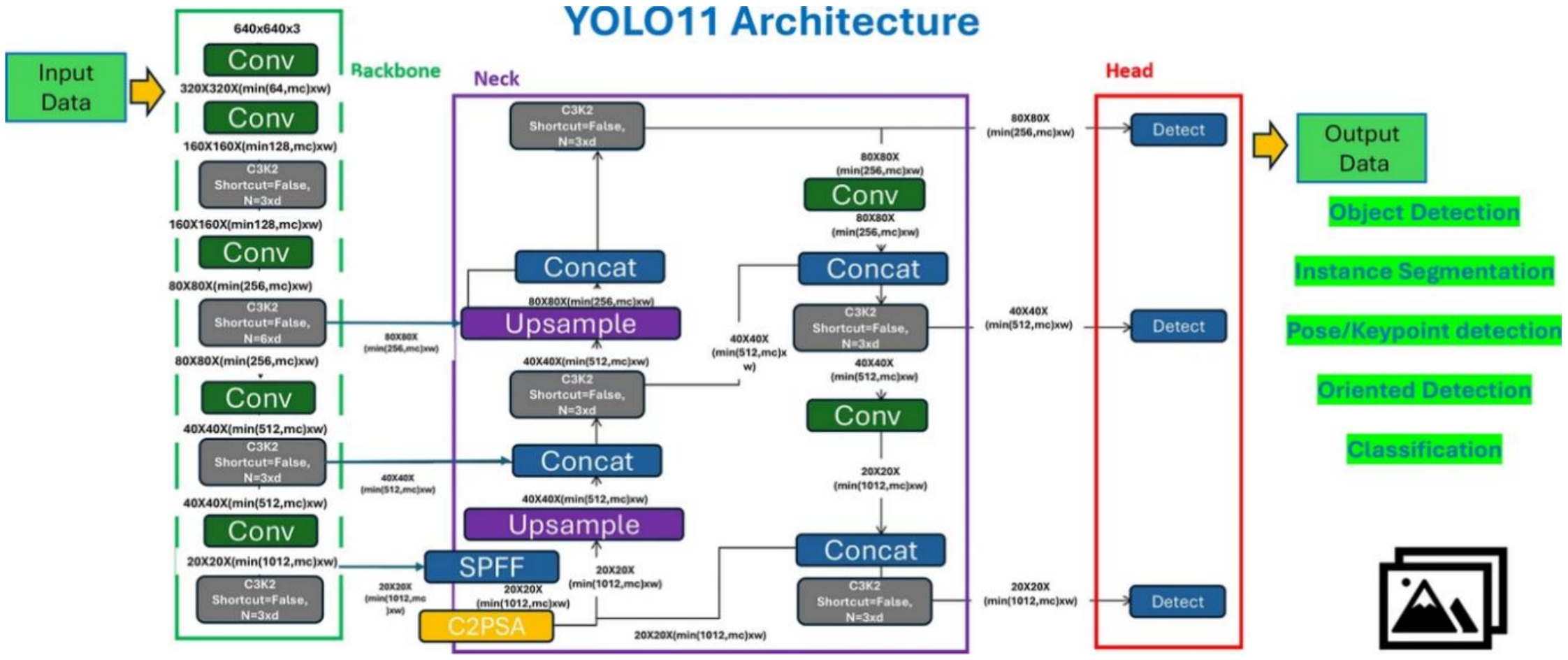

**Figure 2.** YOLOv11 architecture showing the backbone–neck–head pipeline for multi-scale feature extraction and prediction across detection, segmentation, pose estimation, and classification tasks. Reprinted from Sapkota *et al*.[47]

### 2.3. Evaluation metrics and statistical analysis

The performance of the prepared DL model was evaluated using precision, recall, F1-score, area under the receiver operating characteristic curve (AUROC), mean average precision (mAP) at multiple IoU thresholds (mAP50–95), and confusion matrices. These metrics were collectively applied to assess both the WBS and WC tasks. Precision (**Equation 1**) measures the proportion of correctly identified positive predictions among all predicted positives, while recall (**Equation 2**) captures the proportion of actual positive samples correctly detected by the model. The F1-score (**Equation 3**), as the harmonic mean of precision and recall, provides a balanced evaluation by jointly accounting for false positives and false negatives. AUROC reflects the model's ability to discriminate between classes across various threshold settings and is derived from the ROC curve, which plots the true positive rate against the false positive rate. For mask segmentation evaluation, mAP50–95 (**Equation 4**) offers a comprehensive measure by averaging the precision over multiple IoU thresholds. Confusion matrices were also used to visualize the class-wise distribution of predictions and misclassifications.

$$\text{Precision} = \frac{\text{True positives (TP)}}{\text{True positives (TP)} + \text{False positives (FP)}} \tag{1}$$

$$\text{Recall} = \frac{\text{True positives (TP)}}{\text{True positives (TP)} + \text{False negatives (FN)}} \tag{2}$$

$$F1 = 2 \times \frac{\text{Precision} \times \text{Recall}}{\text{Precision} + \text{Recall}} \tag{3}$$

$$\text{mAP}_{50\text{–}95} = \frac{1}{10}\sum_{j=0}^{9} \text{AP}_{\text{IoU}}(0.5 + 0.05j) \tag{4}$$

All reported metrics were calculated using five-fold cross-validation, as illustrated in Figure 3. In this approach, the dataset was divided into five equally sized subsets, known as folds, while maintaining the same class distribution across each subset for WC (stratification). In each cross-validation run, one fold served as the test split (i.e., the portion of data used for validation), while the remaining four folds were combined to form the training split used to train the model. This process was repeated five times so that each fold served once as the test set. For each metric, five individual scores were obtained, one per fold, which were then used to compute the mean and standard deviation to summarize model performance and its variability across different data splits. The mean (μ) and standard deviation (σ) for each metric *M* across *k* folds were calculated using **Equation 5** and **Equation 6**, respectively.

$$\mu = \frac{1}{k}\sum_{i=1}^{k} M_i \tag{5}$$

$$\sigma = \sqrt{\frac{1}{k-1}\sum_{i=1}^{k}(M_i - \mu)^2} \quad (6)$$

where $M_i$ represents the metric value for the i-th fold, and $k = 5$ is the number of folds in cross-validation.

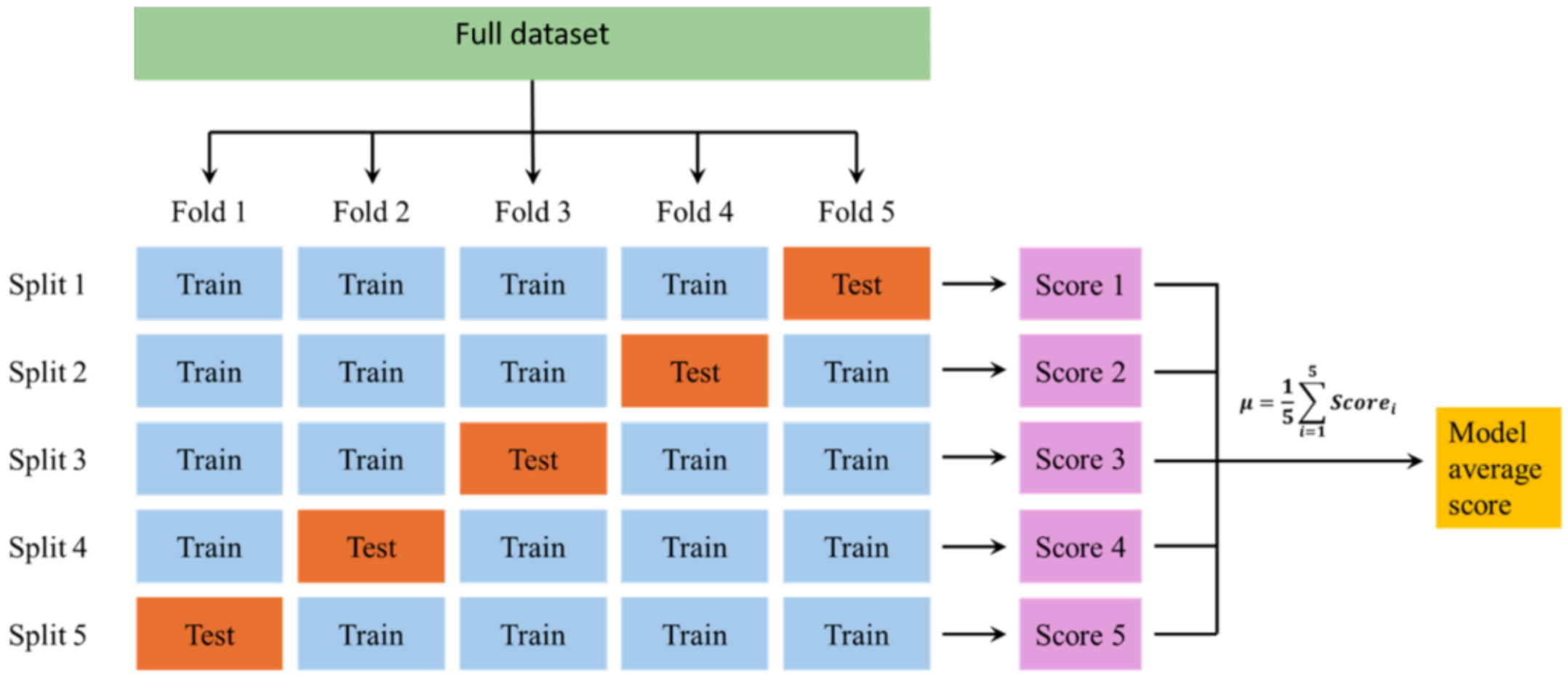

**Figure 3.** Five-fold cross-validation process where each fold serves once as the test set to compute the model's average performance score.

## 3. Results and discussion

As described in **Section 2.3**, we used *k*-fold and stratified *k*-fold cross-validation ($k = 5$) to partition the WBS and WC datasets, respectively. Table 2 summarizes the distribution of wound-class instances across the five folds, confirming consistent representation and balanced sample sizes essential for fair model evaluation. The dataset includes 3,837 wound instances from 2,963 wound images. To further characterize the spatial and dimensional properties of annotated wounds, we analyzed the bounding box distributions across the annotated image pool. Figure 4a presents a two-dimensional histogram of bounding box center coordinates, showing a strong concentration near the image center. This center bias likely arises from standardized clinical photography and implies that spatial augmentations should preserve central positioning to avoid introducing artifacts. Figure 4b illustrates the distribution of bounding box widths and heights (normalized to image dimensions). Most instances fall within the lower-left quadrant, indicating relatively small wound areas with moderate aspect-ratio variability. These findings emphasize the need for multi-scale feature extraction tailored to small-to-medium wound sizes.

Collectively, the analysis reveals structured variability in wound geometry and positioning, informing the architectural choices and training strategies for robust detection.

**Table 2.** Distribution of training and validation instances across five folds for wound boundary segmentation and wound classification tasks

| Task | Class | Split | Fold 1 | Fold 2 | Fold 3 | Fold 4 | Fold 5 |
|---|---|---|---|---|---|---|---|
| Wound boundary segmentation | Wound | Train | 3,088 | 3,055 | 3,098 | 3,063 | 3,044 |
| | | Valid | 749 | 782 | 739 | 774 | 793 |
| Wound classification | Burn injuries | Train | 748 | 744 | 710 | 723 | 727 |
| | | Valid | 165 | 169 | 203 | 190 | 186 |
| | Diabetic foot ulcers | Train | 513 | 500 | 499 | 508 | 512 |
| | | Valid | 120 | 133 | 134 | 125 | 121 |
| | Pressure injuries | Train | 697 | 680 | 702 | 691 | 690 |
| | | Valid | 168 | 185 | 163 | 174 | 175 |
| | Surgical wounds | Train | 468 | 473 | 461 | 473 | 469 |
| | | Valid | 118 | 113 | 125 | 113 | 117 |
| | Vascular ulcers | Train | 682 | 677 | 669 | 663 | 669 |
| | | Valid | 158 | 163 | 171 | 177 | 171 |

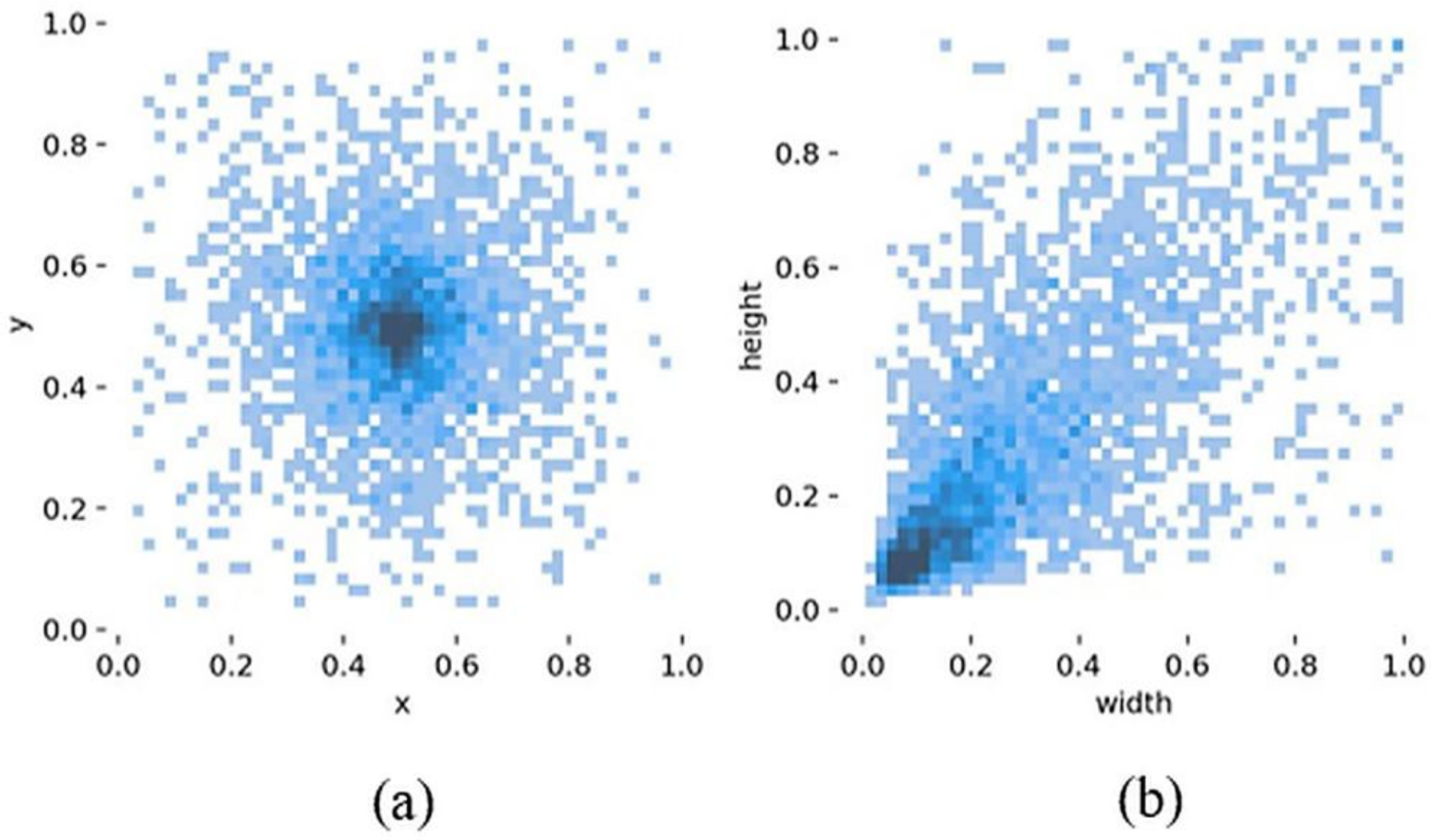

**Figure 4.** Bounding box statistics of wounds. (a) Two-dimensional (2D) histogram of center coordinates (x, y); and (b) 2D histogram of width and height.

To ensure reliable performance, we first examined the dataset for potential bias and class imbalance during preprocessing. As a first ablation, we compared the YOLOv8 and YOLOv11 backbones under identical conditions to determine the more suitable architecture for our task WBS, using the dataset summarized in Table 2. As this comparison focused on selecting the backbone architecture, we employed the lightweight "nano" variants—YOLOv8n and YOLOv11n—to minimize computational cost. Model performance was assessed based on normalized accurate predictions computed from confusion matrices across five-fold cross-validation. As shown in Table 3, YOLOv11n consistently outperformed YOLOv8n, achieving a mean normalized accuracy of 0.88 ± 0.02 compared to 0.87 ± 0.02 for YOLOv8n. This consistent improvement led to selecting YOLOv11 as the primary architecture for the remainder of the study. For the WC task using YOLOv11n, class-wise normalized accurate predictions varied across wound types. DFU and SW achieved the highest scores (0.91 ± 0.04 and 0.91 ± 0.03, respectively), indicating strong feature separability. PI and VU also performed well, with values of 0.83 ± 0.02 and 0.88 ± 0.03. However, BI showed the lowest performance at 0.70 ± 0.03, with approximately 34% of BI samples misclassified as background. This suggests the model struggled to detect BI, potentially because of ambiguous boundaries or low contrast with surrounding skin. Overall, the YOLOv11n model achieved a precision of 0.8936, recall of 0.8478, and mAP50 of 0.9183 for WBS. For WC, the model attained a precision of 0.8701, recall of 0.8216, and mAP50 of 0.8791.

**Table 3.** Normalized accurate prediction scores obtained from confusion matrices for wound boundary segmentation and five-class wound classification tasks across five folds

| Variables | Fold 1 | Fold 2 | Fold 3 | Fold 4 | Fold 5 | Mean ± standard deviation |
|---|---|---|---|---|---|---|
| Wound boundary segmentation | | | | | | |
| Model | | | | | | |
| YOLOv8n | 0.87 | 0.88 | 0.84 | 0.88 | 0.86 | 0.87 ± 0.02 |
| YOLOv11n | 0.9 | 0.87 | 0.88 | 0.88 | 0.85 | 0.88 ± 0.02 |
| Wound classification (YOLOv11n), class ($N = 5$) | | | | | | |
| BI | 0.7 | 0.74 | 0.67 | 0.67 | 0.7 | 0.70 ± 0.03 |
| DFU | 0.87 | 0.89 | 0.88 | 0.94 | 0.95 | 0.91 ± 0.04 |
| PI | 0.85 | 0.82 | 0.85 | 0.82 | 0.81 | 0.83 ± 0.02 |
| SW | 0.88 | 0.94 | 0.9 | 0.96 | 0.89 | 0.91 ± 0.03 |
| VU | 0.91 | 0.85 | 0.9 | 0.86 | 0.87 | 0.88 ± 0.03 |

Abbreviations: BI: Burn injury; DFU: Diabetic foot ulcer; PI: Pressure injury; SW: Surgical wound; VU: Vascular ulcer.

To investigate the impact of data augmentation on BI classification and overall model performance, we applied augmentation techniques to increase training data diversity, enhance model robustness, and improve the detection of subtle features—particularly in BI cases with low contrast or indistinct wound boundaries. Each task-specific dataset contained 2,963 original images and 7,147 total images after augmentation using the strategies detailed in **Section 2.1**. Validation images were kept unchanged for both non-augmented and augmented experiments to ensure fair comparison. Table 4 summarizes the performance of five YOLOv11 variants (n, s, m, l, x) on the augmented dataset. Data augmentation substantially improved YOLOv11n's performance across both WBS and WC tasks. For WBS, precision improved from 0.8936 to 0.9263, recall from 0.8478 to 0.9307, and mAP50 from 0.9183 to 0.9587. In WC, precision increased from 0.8701 to 0.8808, recall from 0.8216 to 0.8415, and mAP50 from 0.8791 to 0.9024. These gains are attributed to the augmented dataset's increased variability, which enabled the model to learn more robust, generalizable representations—especially for visually ambiguous wounds such as BI. Augmentations including changes in rotation, illumination, and background context reduced overfitting and improved generalization, resulting in better validation performance. In WBS, performance improved with model size, with YOLOv11x achieving the highest F1-score of 0.9341 and mAP50 of 0.9629, while the lightweight YOLOv11n remained competitive with an F1-score of 0.9285. Every model showed similar performance for WC. YOLOv11m led with an mAP50 of 0.9194, closely followed by YOLOv11l and YOLOv11x. Although YOLOv11x did not top the WC task, it maintained strong performance across both tasks, highlighting its versatility. To characterize the computational requirements of each variant, Table 5 reports the parameter count, giga floating-point operations per second (GFLOPs), model size, and inference speed for both tasks. Because the WBS and WC models share the same architecture, their parameters, GFLOPs, and model size were identical and listed once per variant, while latency and throughput are reported separately for each task. Computational cost scaled with model size, from YOLOv11n (2.84 M parameters, 10.4 GFLOPs, 6.0 MB) to YOLOv11x (62.05 M parameters, 319.7 GFLOPs, 249.1 MB). Inference was measured on an NVIDIA V100 graphics processing unit at batch size 1 to reflect single-image, point-of-care use. All five variants operated in real time, from approximately 128 FPS for YOLOv11n to 33 FPS for YOLOv11x, with WBS and WC showing nearly identical speed at each model size. These results indicate that the proposed models, particularly the lightweight variants, are computationally efficient enough for deployment on resource-constrained hardware.

**Table 4.** Performance comparison of YOLOv11 model variants on wound boundary segmentation and classification using the augmented dataset

| Models | Precision | Recall | F1-score | mAP50 | mAP50–95 |
|---|---|---|---|---|---|
| Wound boundary segmentation | | | | | |
| YOLOv11n | 0.9263 | 0.9307 | 0.9285 | 0.9587 | 0.7135 |
| YOLOv11s | 0.9271 | 0.9269 | 0.9269 | 0.9557 | 0.7251 |
| YOLOv11m | 0.9201 | 0.9335 | 0.9267 | 0.9584 | 0.7355 |
| YOLOv11l | 0.9306 | 0.9341 | 0.9323 | 0.9586 | 0.7361 |
| YOLOv11x | 0.9373 | 0.9310 | 0.9341 | 0.9629 | 0.7399 |
| Wound classification | | | | | |
| YOLOv11n | 0.8808 | 0.8415 | 0.8607 | 0.9024 | 0.6657 |
| YOLOv11s | 0.8822 | 0.8583 | 0.8701 | 0.9136 | 0.6851 |
| YOLOv11m | 0.9045 | 0.8447 | 0.8736 | 0.9194 | 0.6950 |
| YOLOv11l | 0.8976 | 0.8625 | 0.8797 | 0.9171 | 0.6936 |
| YOLOv11x | 0.8857 | 0.8679 | 0.8767 | 0.9116 | 0.6923 |

Abbreviation: mAP: Mean average precision.

**Table 5.** Computational characteristics of the five YOLOv11 variants

| Variant | Parameters (M) | GFLOPs | Size (MB) | Latency (ms) | | FPS | |
|---|---|---|---|---|---|---|---|
| | | | | WBS | WC | WBS | WC |
| YOLOv11n | 2.84 | 10.4 | 6.0 | 7.83 | 7.99 | 127.7 | 125.2 |
| YOLOv11s | 10.08 | 35.6 | 20.5 | 8.35 | 7.92 | 119.8 | 126.3 |
| YOLOv11m | 22.36 | 123.6 | 45.2 | 13.04 | 13.10 | 76.7 | 76.3 |
| YOLOv11l | 27.62 | 142.7 | 55.9 | 15.74 | 15.74 | 63.5 | 63.5 |
| YOLOv11x | 62.05 | 319.7 | 249.1 | 30.38 | 30.50 | 32.9 | 32.8 |

Note: Parameters, GFLOPs, and model size are architecture-dependent and identical across tasks; latency and FPS were measured separately for the WBS and WC models on an NVIDIA V100 GPU at batch size 1.

Abbreviations: FPS: Frames per second; GFLOPs: Giga floating-point operations per second; WC: Wound classification; WBS: Wound boundary segmentation.

Figure 5 provides a detailed analysis of the YOLOv11m model's training behavior and evaluation performance across 150 epochs for the WC task. As shown in Figure 5a, the training and validation loss curves show consistent, stable convergence across all major loss components. The box loss declined smoothly and gradually in both phases, indicating improved localization of wound boundaries. The segmentation loss decreased substantially, reflecting improved accuracy in generating instance-level wound masks. A notable drop in classification loss demonstrates improved model capability in differentiating wound types. Similarly, the distribution focal loss, which supports precise bounding-box regression, steadily declines, confirming the model's increasing localization accuracy. Collectively, these trends underscore stable convergence and effective joint optimization across all learning objectives. Figure 5b illustrates the progression of validation metrics for bounding-box (B) and segmentation-mask (M) predictions. Precision and recall generally increased during training, indicating improved instance detection, localization, classification, and mask delineation. Similarly, mAP50 and mAP50–95 increased before approaching a plateau, demonstrating improved performance across multiple IoU thresholds. Overall, these trends indicate effective learning and increasingly accurate bounding-box and segmentation-mask predictions on the validation data.

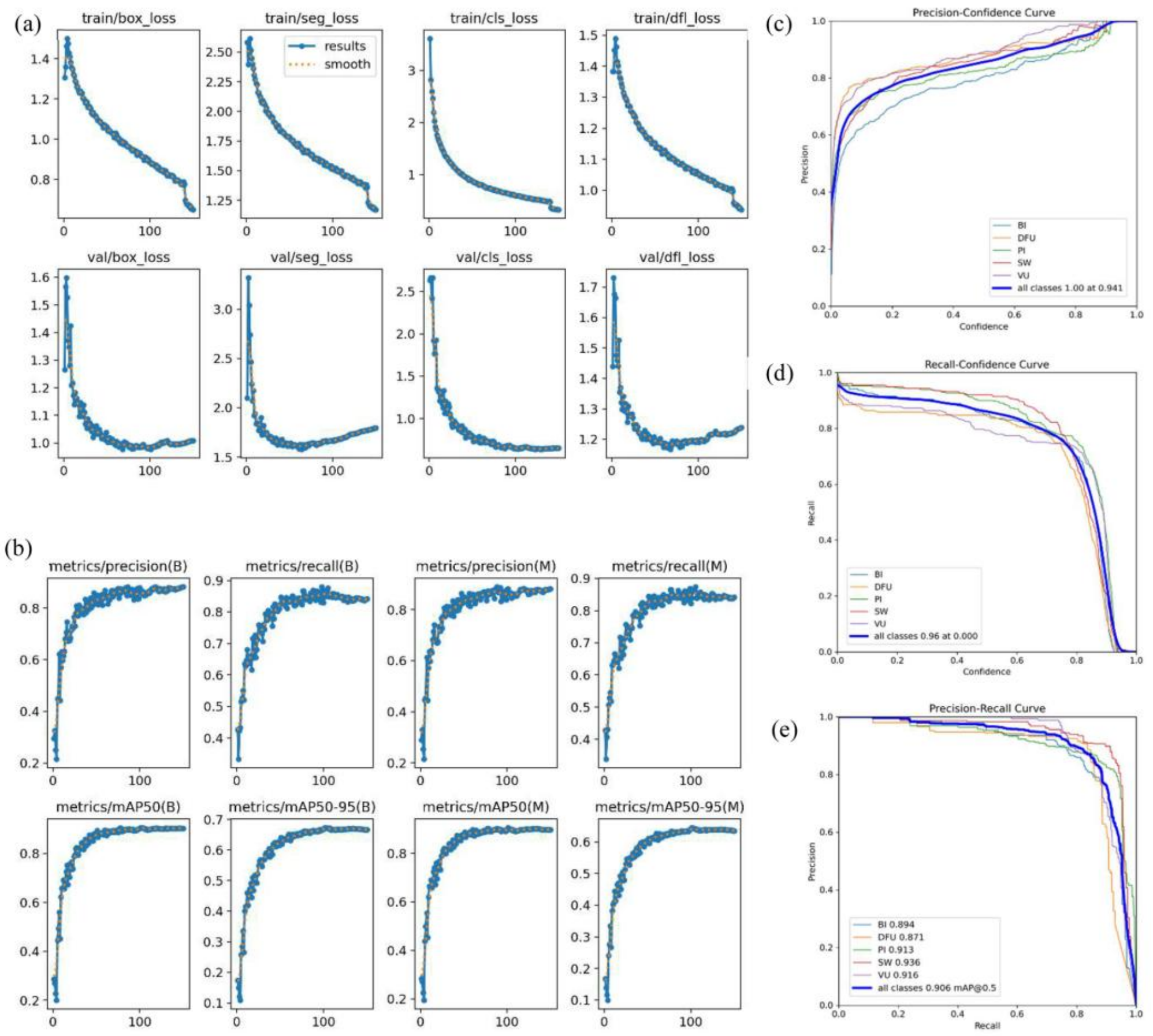

**Figure 5.** Visualization of the YOLOv11m model's training and evaluation for wound classification: (a) training and validation loss curves, (b) performance metrics, (c) precision–confidence curve, (d) recall–confidence curve, and (e) precision–recall curve. Abbreviations: BI: Burn injury; DFU: Diabetic foot ulcer; mAP: Mean average precision; PI: Pressure injury; SW: Surgical wound; VU: Vascular ulcer.

Figure 5c–e collectively provide deeper insights into the model's confidence calibration and predictive reliability. The precision-confidence curve in Figure 5c shows that high-confidence predictions are typically accurate, resulting in fewer false positives. The recall–confidence curve in Figure 5d indicates that recall declined gradually as confidence thresholds increased because lower-confidence true positives were excluded. However, the model retained a large proportion of correct predictions even at higher confidence levels. Together, these curves suggest a well-calibrated confidence distribution, with the model maintaining a favorable balance between precision and recall. Supporting this, the precision-recall curve in Figure 5e exhibits a high-curvature profile, indicating reliable classification performance across thresholds. Overall, these plots demonstrate that the YOLOv11m model makes accurate predictions while assigning interpretable confidence scores—an important feature for applications where decision-making may be guided by the certainty of model outputs.

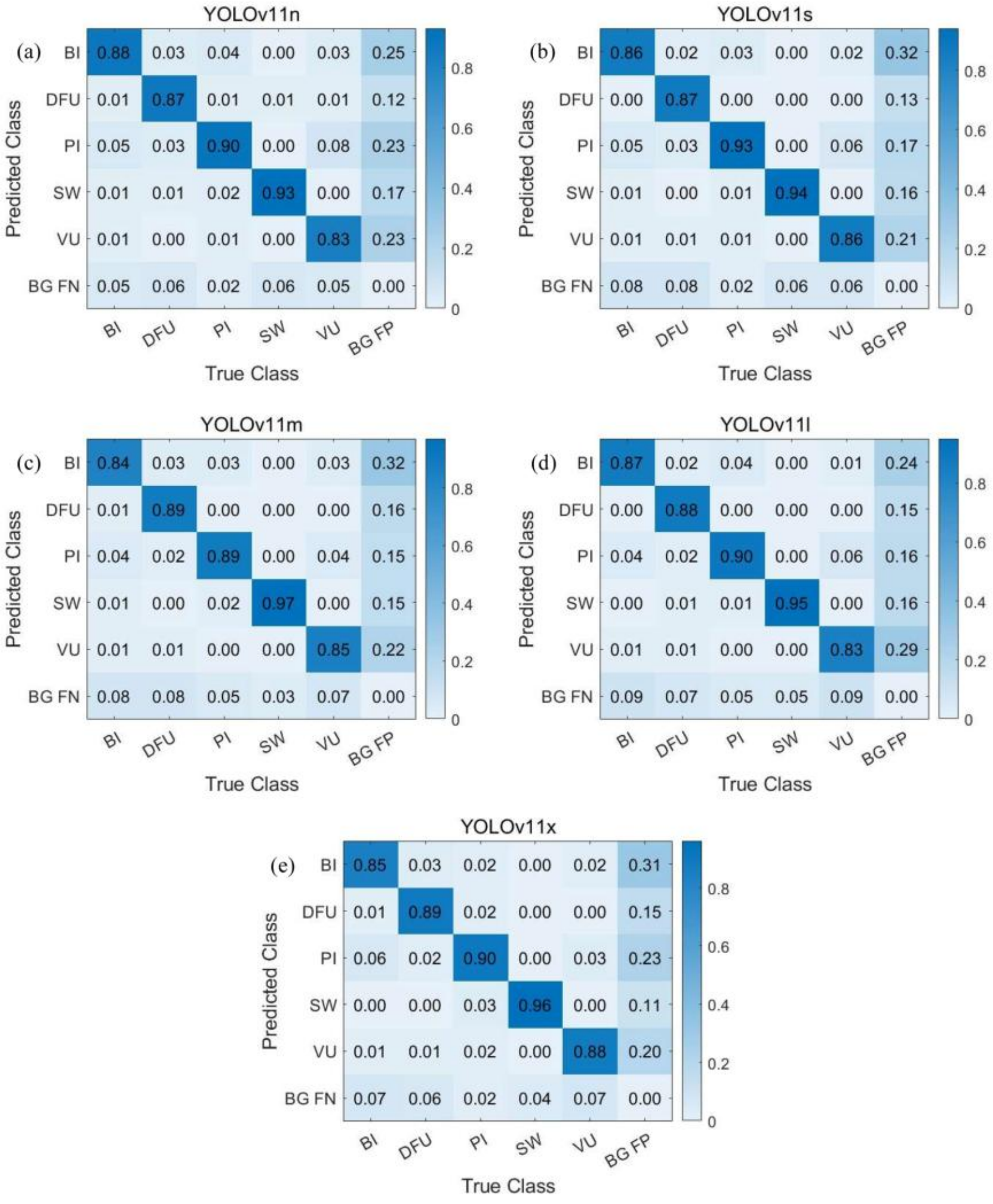

**Figure 6.** Confusion matrices of five YOLOv11 model variants for multi-class wound classification across BI, DFU, PI, SW, and VU categories. (a) YOLOv11n. (b) YOLOv11s. (c) YOLOv11m. (d) YOLOv11l. (e) YOLOv11x. Abbreviations: BG: Background; BI: Burn injury; DFU: Diabetic foot ulcer; FN: False negative; FP: False positive; PI: Pressure injury; SW: Surgical wound; VU: Vascular ulcer.

Figure 6 presents the normalized confusion matrices for five YOLOv11 model variants (n, s, m, l, x), highlighting their performance in WC. In each matrix, the horizontal axis represents the true wound classes—BI, DFU, PI, SW, and VU—while the vertical axis shows the predicted classes. Additional columns and rows labeled background false positives and background false negatives quantified the model's errors in misidentifying background as a wound or failing to detect actual wounds, respectively. The diagonal elements indicate correct classifications, whereas off-diagonal values represent misclassifications between wound types. To quantify performance, we computed the mean and standard deviation of normalized class-wise accurate predictions across the five wound types for each model. All variants achieved comparable results, with average normalized predictions ranging from 0.882 (YOLOv11n) to 0.896 (YOLOv11x), suggesting that each variant can robustly detect WC. YOLOv11x achieved the highest mean normalized accuracy (0.896 ± 0.041), though its advantage over YOLOv11s (0.892 ± 0.036) and YOLOv11m (0.888 ± 0.048) was marginal. YOLOv11l (0.886) and YOLOv11n (0.882) also demonstrated strong performance, reinforcing the reliability of even lightweight models. Class-wise analysis showed consistently high normalized accurate predictions for SW and PI across all variants. In contrast, BI and VU posed greater WC challenges, as reflected in their comparatively lower normalized predictions and elevated background false-positive rates. This suggests that BI and VU, particularly in low-severity stages, closely resemble normal skin, making them harder to distinguish. Since our dataset includes wound images across a wide severity spectrum, including early-stage wounds with subtle features, these results also demonstrate the model's generalization ability. Notably, most incorrect predictions were attributed to background false negatives rather than interclass confusion, indicating that actual class-to-class misclassifications are relatively low. When misclassifications did occur, they were most prominent between BI and PI, and between PI and DFU. This aligns with our dataset observations, where first- and second-degree BI share visual traits with early-stage PI (stages 1 and 2) and deep tissue injuries. Similarly, overlap between DFU and PI likely stems from shared anatomical locations and visual resemblance in advanced stages (stages 3 and 4). Augmentation strategies helped mitigate these issues by exposing the model to a wider range of subtle wound features during training. SW, by contrast, were detected with consistently high accuracy, likely because of their well-defined edges, cleaner wound beds, wound morphology, and more uniform presentation. These findings suggest that all YOLOv11 variants can deliver high performance with only minor trade-offs between accuracy and model size. Among them, YOLOv11x provided the most balanced and robust results, while YOLOv11m offered an attractive trade-off between accuracy and computational efficiency.

Figure 7 shows qualitative WC results using the YOLOv11x model, with side-by-side comparisons of original images and corresponding detection outputs across five wound types. Across all classes, the model demonstrated high visual accuracy in detecting and classifying wounds of all sizes. Bounding boxes aligned well with wound boundaries and tightly fit the affected regions without significant overreach. Segmentation masks also closely matched the ground-truth annotations. The predicted labels were consistent with clinical expectations, even in challenging scenarios involving variations in lighting, skin tones, and wound severity. In several cases, wounds were partially occluded or situated in visually cluttered contexts (e.g., surgical drapes or textured skin), yet the model correctly identified them without false positives in surrounding areas. Furthermore, the detection performance remained stable across wound types with high intra-class variability. For example, BI examples ranged from blistered skin to deep eschar, while PI and DFUs exhibited varying degrees of tissue necrosis and anatomical positioning. Despite this variability, the model maintained consistent performance, suggesting it learned to generalize effectively from the augmented training dataset. Unlike whole-image classification models, which output only a class label and therefore rely on post-hoc techniques such as gradient-weighted class activation mapping or feature-embedding visualization to approximate the image regions influencing a prediction[55,56], our instance segmentation framework provides this spatial interpretability directly: every prediction is accompanied by a pixel-level mask and bounding box that localize the exact wound region on which the classification is based (Figure 7). This is consistent with prior wound-segmentation studies, which likewise convey model behavior through the segmentation output itself rather than through post-hoc saliency maps.[25] The confusion matrices (Figure 6) and class-wise performance analysis further clarify class-level separability, showing which wound types are reliably distinguished and which remain visually similar (e.g., BI and PI).

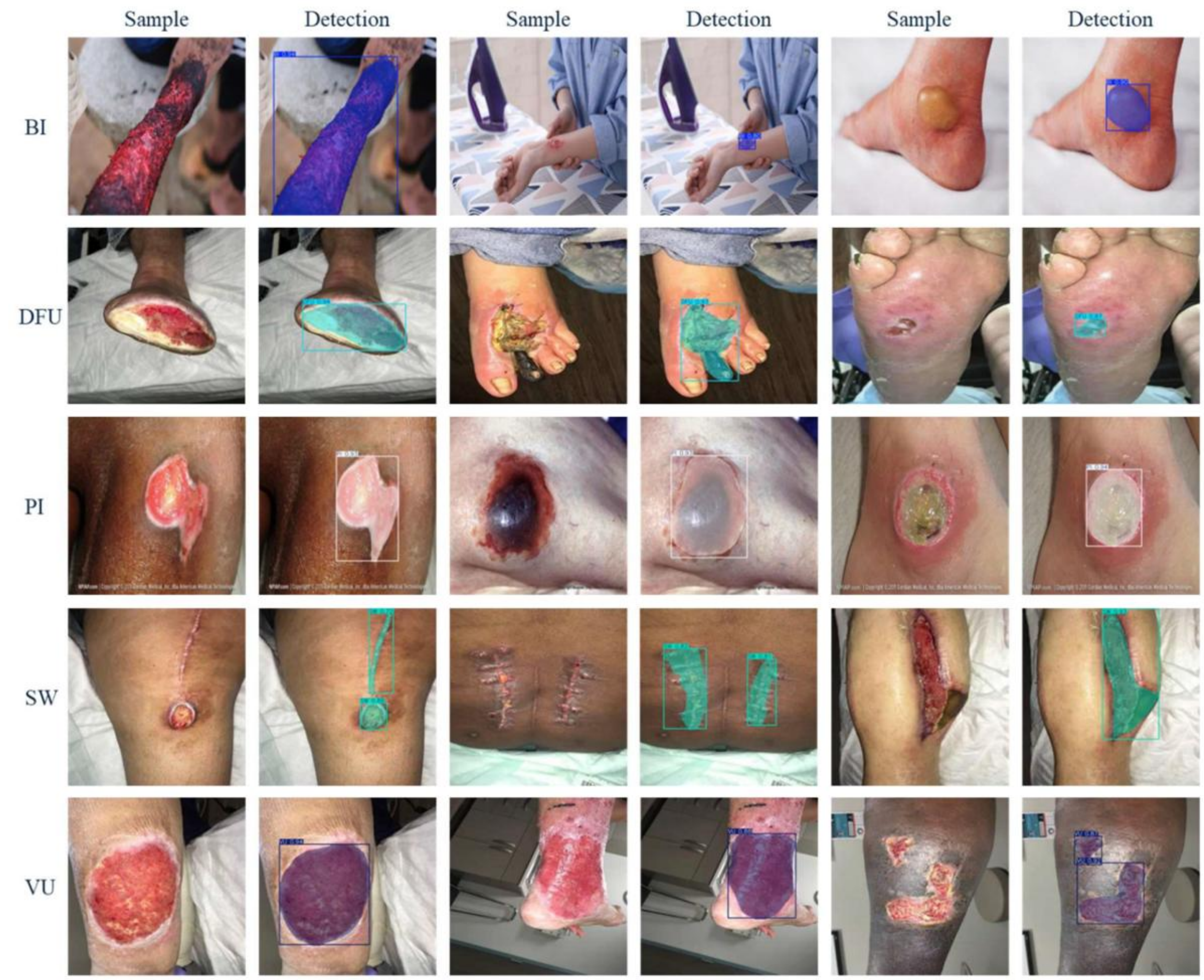

**Figure 7.** Visual examples of wound classification using YOLOv11x showing the original and detection images side by side. Abbreviations: BI: Burn injury; DFU: Diabetic foot ulcer; PI: Pressure injury; SW: Surgical wound; VU: Vascular ulcer.

The performance of the proposed YOLOv11-based WC model was evaluated using ROC curves in a one-vs.-rest manner for each wound type as shown in Figure 8. As the threshold decreased, the model became more permissive, increasing the true positive rate by capturing more actual positives but at the cost of a higher false positive rate due to more false alarms. The ROC curves for all classes showed a sharp rise toward the top-left corner, demonstrating the model's ability to maintain high performance across varying thresholds. The model achieved consistently high AUC scores across all classes, with BI, VU, DFU, and PI each obtaining an AUC of 0.96 or higher, while SW achieved a perfect AUC of 1.00. These results indicate the model's strong discriminative capability in correctly identifying wound types while minimizing false positives.

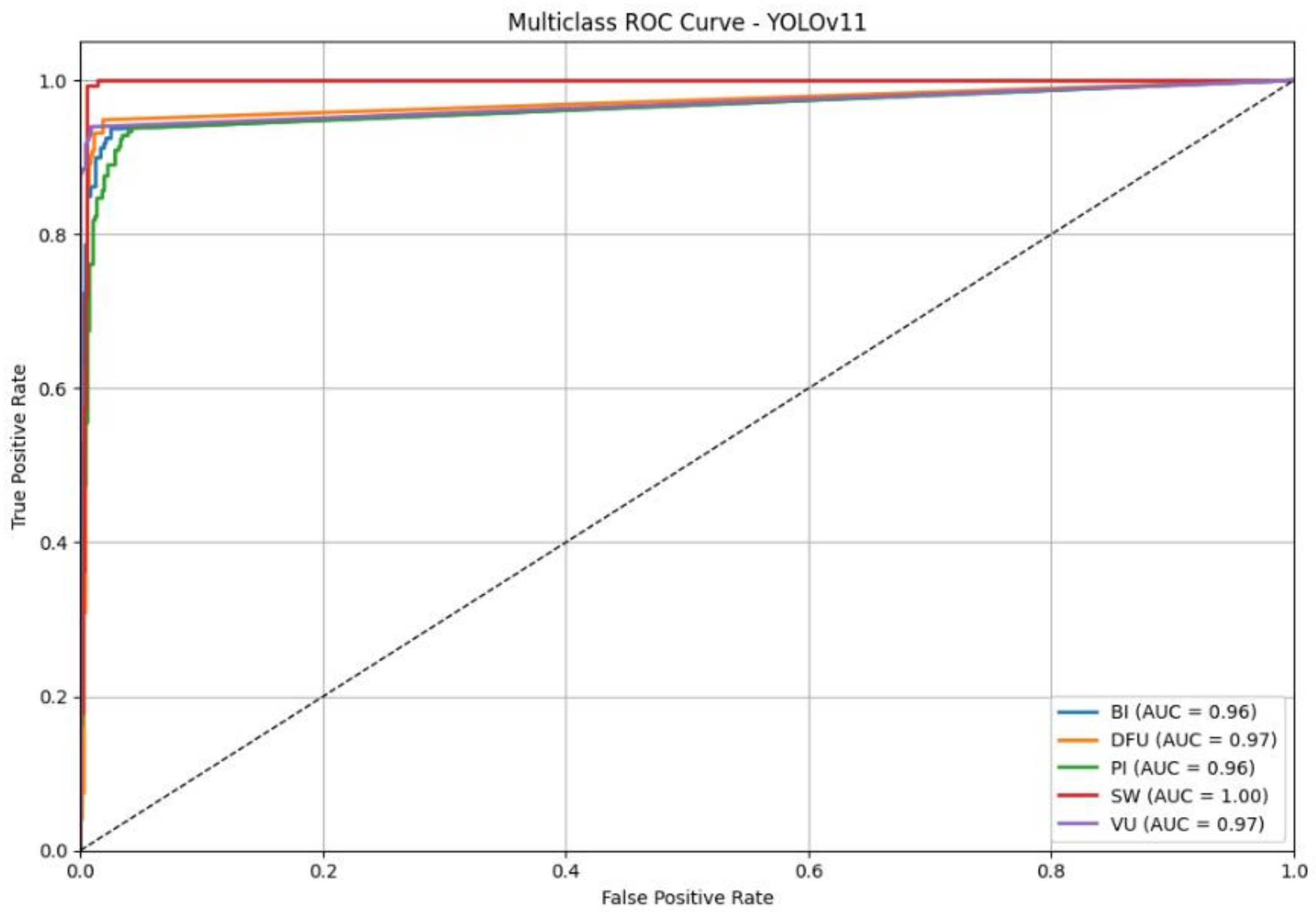

**Figure 8.** Multi-class receiver operating characteristic curve of the YOLOv11x model for wound classification, illustrating the trade-off between true positive rate and false positive rate for each wound type. Abbreviations: AUC: Area under the curve; BI: Burn injury; DFU: Diabetic foot ulcer; PI: Pressure injury; SW: Surgical wound; VU: Vascular ulcer.

To contextualize the performance of our proposed YOLOv11-based system, we compared it against several widely used classification and instance segmentation models, as shown in **Table 6**. Among the classification models, YOLOv11 outperformed traditional architectures such as VGG16, AlexNet, and ResNet101 across all metrics for the same dataset. It achieved the highest F1-score of 0.8767, compared to 0.8732 (VGG16), 0.8401 (AlexNet), and 0.8020 (ResNet101). Our model was trained on all major wound types, such as BI, DFU, PI, SW, and VU, whereas other models were trained on fewer wound classes. Anisuzzaman *et al*.[35] developed a multi-modal classifier that was trained on DFU, PI, SW, and VU (without BI). They showed that when the number of wound classes in the training model increases, performance metrics such as accuracy decrease.[35] Patel *et al*.[34] performed whole-image classification for DFU, PI, SW, and VU using the AZH dataset. Their model achieved F1-scores ranging from 0.8282 to 1.0000 for binary classification (2 classes). When extended to three classes, the F1-scores ranged from 0.7487 to 0.9254, and for the four-class model, the F1-scores varied between 0.7675 and 0.8226.[34] Our model achieved an F1-score of 0.8767 with five classes, surpassing the previous models. The classifier models did not generate segmentation masks, whereas the YOLOv11 model provided both classification results and corresponding segmentation masks. This indicates YOLOv11's superior ability to not only identify wound classes accurately but also to minimize both false positives and false negatives more effectively than the baseline classifiers. Similarly, our current YOLOv11 model had a higher F1-score than the previous instance segmentation models such as DeepLabV3, UNet, and Mask R-CNN.[9,41]

**Table 6.** Performance comparison of classification and instance segmentation models based on precision, recall, F1-score, and key architectural features

| Type | Model | Precision | Recall | F1 | Features |
|---|---|---|---|---|---|
| Classification | VGG16 (current) | 0.8739 | 0.8737 | 0.8732 | Image-level classification |
| | AlexNet (current) | 0.8432 | 0.8393 | 0.8401 | Bounding boxes (optional) |
| | ResNet101 (current) | 0.8112 | 0.7991 | 0.8020 | Low computational cost |
| Instance segmentation | YOLOv11 (current) | 0.8857 | 0.8679 | 0.8767 | Image classification |
| | DeeplabV3[9] | 0.7533 | 0.9082 | 0.8235 | Object detection with bounding boxes |
| | UNet[9] | 0.7722 | 0.8453 | 0.8071 | Pixel-level segmentation masks |
| | Mask R-CNN[41] | 0.77 | 0.72 | 0.75 | Requires moderate to heavy computational power |

While the proposed YOLOv11-based model demonstrated strong performance in both WBS and WC, several limitations remain. Our dataset is large, but more data can be acquired to improve performance and reliability. More images with variable skin tones, ages, sexes, and body locations can be included. Aggregating images from heterogeneous public repositories may also introduce selection and source biases, such as uneven representation of wound presentations and imaging conditions; we mitigated this through quality filtering, duplicate removal, balanced class representation across folds (Table 2), and a single consistent re-annotation protocol, though residual bias cannot be fully excluded. In addition, generalization in this study was assessed through internal five-fold cross-validation rather than validation on an independent external dataset, and the models have not yet undergone prospective evaluation in a real-world clinical workflow; both external and prospective validation are necessary before clinical deployment. Access to clinical data is sometimes limited; therefore, a generative adversarial network for data generation should be explored. The current model does not incorporate clinical metadata such as wound location, patient history, or comorbidities, which could provide valuable context and enhance WC accuracy—especially in visually ambiguous or overlapping wound categories such as DFU and PI. Manual annotation of segmentation masks introduces variability and is resource-intensive. Semi-supervised learning can reduce annotation time and workload by automating the annotation process. We used pretrained YOLOv11 segmentation models trained on the COCO dataset.[57] However, domain-specific transfer learning can speed up model training and improve accuracy. Although YOLOv11 offers real-time detection capability, the model has not yet been tested or optimized for deployment on resource-constrained POC devices, which is essential for bedside or remote clinical use.

## 4. Conclusion

This study presented a DL system based on the YOLOv11 architecture for real-time WBS and multi-class WC across five major wound types: BI, PI, DFU, VU, and SW. Using a balanced dataset of 2,963 expert-annotated images with extensive augmentations, the model was trained and validated through five-fold cross-validation to ensure reliability and generalizability. The low variation across folds (±0.02 for WBS and <±0.04 for WC) on YOLOv11n demonstrates consistent performance, indicating that the model's predictions are robust and not dependent on specific data subsets.

Five YOLOv11 variants n, s, m, l, and x were tested for performance in performing WBS and WC. Among the tested variants, YOLOv11x achieved the highest performance, with an F1-score of 0.9341 for WBS. For WC, YOLOv11m showed the best performance when overall performance and computational cost were considered. The lightweight YOLOv11n variant also delivered competitive results (F1 = 0.9285 for WBS, 0.8607 for WC) while using 25% of YOLOv11x's inference latency, underscoring its suitability for point-of-care or mobile deployments. Data augmentation significantly enhanced the model's robustness, particularly in distinguishing low-contrast BI, with all performance metrics showing improvement over the non-augmented dataset. For WBS with YOLOv11n, precision improved from 0.8936 to 0.9263, recall from 0.8478 to 0.9307, and mAP50 from 0.9183 to 0.9587. In WC with YOLOv11n, precision increased from 0.8701 to 0.8808, recall from 0.8216 to 0.8415, and mAP50 from 0.8791 to 0.9024.

Comparative evaluations showed that the YOLOv11-based model outperformed conventional models such as VGG16, ResNet101, and Mask R-CNN, achieving higher accuracy while also generating instance-level segmentation masks. The model effectively reduced false positives and maintained strong discriminative capability, achieving AUC scores ≥ 0.96 for all classes, including visually subtle BI and VU. Qualitative visualizations confirmed that the system produced well-aligned wound boundaries and consistent predictions under varied lighting and background conditions.

In summary, the YOLOv11-based fine-tuned DL model demonstrates accurate, real-time wound detection, segmentation, and classification across diverse wound categories. These findings establish a pathway to integrate AI-powered wound assessment tools into clinical and remote monitoring workflows. Future work should expand dataset diversity across skin tones, ages, sexes, and anatomical location; explore generative models to address limited clinical data; incorporate multi-modal clinical metadata; adopt semi-supervised annotation and domain-specific transfer learning; and optimize the model for deployment on resource-constrained point-of-care devices.

## Acknowledgments

The authors acknowledge the Missouri University of Science and Technology for providing access to the high-performance computing and software resources used in this study. The authors also acknowledge the publicly available image repositories that supported dataset development.

## Funding

None.

## Conflict of interest

The authors declare no competing interests.

## Author contributions

*Conceptualization:* Mehedi Hasan Tusar, Fateme Fayyazbakhsh

*Data curation:* Mehedi Hasan Tusar, Fateme Fayyazbakhsh, Igor Melnychuk

*Formal analysis:* Mehedi Hasan Tusar

*Methodology:* Mehedi Hasan Tusar, Fateme Fayyazbakhsh

*Project administration:* Ming C. Leu

*Software:* Mehedi Hasan Tusar

*Supervision:* Fateme Fayyazbakhsh, Igor Melnychuk, Ming C. Leu

*Validation:* Fateme Fayyazbakhsh, Igor Melnychuk

*Visualization:* Mehedi Hasan Tusar, Fateme Fayyazbakhsh

*Writing – original draft:* Mehedi Hasan Tusar

*Writing – review & editing:* Fateme Fayyazbakhsh, Igor Melnychuk, Ming C. Leu

All authors read and approved the final manuscript.

## Ethics approval and consent to participate

Not applicable. This study used publicly available, de-identified wound images obtained from open-access databases and online repositories. No direct patient contact, intervention, or access to identifiable private health information was involved. Therefore, institutional review board approval and informed consent to participate were not required.

## Consent to publication

Not applicable.

## Availability of data

The datasets supporting the findings of this study are publicly available in Roboflow under the project identifiers wound5class and wbs-da6m0. Trained models and additional implementation details may be made available from the corresponding author upon reasonable request.

## Further disclosure